\documentclass{article} % For LaTeX2e
\usepackage{iclr2015,times}
\usepackage{hyperref}
\usepackage{url}
\usepackage{epsfig}
\usepackage{graphicx}
\usepackage{tabularx}
\usepackage{array}
\usepackage{mathtools}

\DeclarePairedDelimiter\floor{\lfloor}{\rfloor}

\graphicspath{{./images/}}

\title{Striving for Simplicity: \\ The All Convolutional Net} 
\author{
Jost Tobias Springenberg\footnotemark[1] \,, Alexey
Dosovitskiy\thanks{Both authors contributed equally to this work.}\, , Thomas Brox, Martin Riedmiller \\
Department of Computer Science\\
University of Freiburg\\
Freiburg, 79110, Germany \\
\texttt{\{springj, dosovits, brox, riedmiller\}@cs.uni-freiburg.de} \\
}

\iclrfinalcopy % Uncomment for camera-ready version

\begin{document}

\maketitle

\begin{abstract}
Most modern convolutional neural networks (CNNs) used for
object recognition are built using the same principles: Alternating convolution and max-pooling
layers followed by a small number of fully connected layers.
We re-evaluate the state of the art for object recognition from small
images with convolutional networks, questioning the necessity of
different components in the pipeline. We find that
max-pooling can simply be replaced by a convolutional layer with increased stride
without loss in accuracy on several image recognition
benchmarks. Following this finding -- and building on other recent work
for finding simple network structures -- we propose a new
architecture that consists solely of convolutional layers and yields
competitive or state of the art performance on several object recognition datasets (CIFAR-10, CIFAR-100, ImageNet). 
To analyze the network we introduce a new variant of the ``deconvolution approach'' for visualizing features learned by CNNs, which can be applied to a broader range of network structures than existing approaches. 
%Max-pooling is a component of the vast majority of modern convolutional neural networks. We question if it is at all necessary in modern ConvNets and substitute the max-pooling procedure by simple stride in convolutional layer \TODO{maybe reformulate it depending on what we end up doing}. Quite surprisingly, when experimenting on CIFAR-10 and ImageNet, we found that such simplifications does not reduce the performance of the network, and may even slightly help. Moreover, this simplifi
\end{abstract}

\section{Introduction and Related Work}
The vast majority of modern convolutional neural networks (CNNs) used
for object recognition are built using the same principles: They 
use alternating convolution and max-pooling layers followed by a small number of fully
connected layers~(e.g. \citet{Jarrett_2009,Krizhevsky_NIPS2012,Ciresan_2011}). Within each of these layers piecewise-linear
activation functions are used. The networks are
typically parameterized to be large and regularized during training
using dropout. A considerable amount of research has over the last years 
focused on improving the performance of this basic pipeline. Among
these two major directions can be identified.
First, a plethora of extensions were recently proposed to
enhance networks which follow this basic scheme. Among these the most
notable directions are work on using more complex activation functions
\citep{Goodfellow2013,Lin_2014,SrivastavaSchmid_2013} techniques for improving class inference
\citep{Stollenga_2014,Nitish2013} as well as procedures for improved regularization
\citep{ZeilerStochastic2013,SprRied2014a,WanLi2013} and layer-wise pre-training using label information
\citep{Lee_2014}.
Second, the success of CNNs for large scale object recognition in the
ImageNet challenge \citep{Krizhevsky_NIPS2012} has stimulated research
towards experimenting with the different architectural choices in
CNNs. Most notably the top entries in the 2014 ImageNet challenge
deviated from the standard design principles by either introducing
multiple convolutions in between pooling layers~\citep{VGG_2014} or by
building heterogeneous modules performing convolutions and pooling at multiple
scales in each layer \citep{GoingDeep_2014}.

Since all of these extensions and different architectures come with
their own parameters and training procedures the question arises which
components of CNNs are actually  necessary for achieving
state of the art performance on current object recognition datasets.
We take a first step towards answering this question by
studying the most simple architecture we could conceive: a homogeneous network
solely consisting of convolutional layers, with occasional
dimensionality reduction by using a stride of 2. Surprisingly, we
find that this basic architecture -- trained using vanilla stochastic
gradient descent with momentum -- reaches state of the art
performance without the need
for complicated activation functions, any response normalization or max-pooling. We empirically study
the effect of transitioning from a more standard architecture to our
simplified CNN by performing an ablation study on CIFAR-10 and compare
our model to the state of the art on CIFAR-10, CIFAR-100 and the
ILSVRC-2012 ImageNet dataset. Our results both confirm the effectiveness
of using small convolutional layers as recently proposed by \citet{VGG_2014} and give rise to interesting new
questions about the necessity of pooling in CNNs. Since dimensionality
reduction is performed via strided convolution rather than max-pooling in our
architecture it also naturally lends itself to studying questions about the
invertibility of neural networks \citep{Estrach_2014}. For a first step in
that direction we study properties of our network using a
deconvolutional approach similar to \citet{Zeiler_ECCV2014}.

% This is what Alexey had here before
%Almost everyone almost always used max pooling, starting from (Neocognitron, LeNet, ??). Some people did not, e.g. the DeepMind guys. It actually seems like things worked. We wondered how important is actually the max pooling. Turns out, not so important, you can do just as well without it. Absence of max pooling does not only make the network architecture simpler and more consistent, but also allows to (deconvolve, localize?)

\section{Model description - the all convolutional network}
\label{sect:model}
The models we use in our experiments differ from standard CNNs in
several key aspects. First -- and most interestingly -- we replace
the pooling layers, which are present in practically all modern CNNs used for object
recognition, with standard convolutional layers with stride two. 
 To understand why this procedure can work it helps to recall the standard
 formulation for defining convolution and pooling operations in
 CNNs. 
%We will thus briefly cover it here. 
Let $f$ denote a feature map produced by some layer of a CNN. It can be described as
 a 3-dimensional array of size $W \times H \times N$ where $W$ and $H$ are the width and
 height and $N$ is the number of channels (in case $f$ is the output of a convolutional layer, $N$ is the number of filters in this layer). 
 Then p-norm subsampling (or pooling) 
 with pooling size $k$ (or half-length $k/2$) and stride $r$ applied to the feature map $f$ is a 3-dimensional array $s(f)$ with the following entries:   
 \begin{equation}
   s_{i,j,u}(f) = \left (\sum_{h=-\floor{k/2}}^{\floor{k/2}} \sum_{w=-\floor{k/2}}^{\floor{k/2}}
     |f_{g(h,w,i,j,u)}|^p \right )^{1/p},
  \label{eq:pooling}
 \end{equation}
 where $g(h,w,i,j,u) = (r \cdot i + h, r \cdot j + w, u)$ is the function mapping from
 positions in $s$ to positions in $f$ respecting the stride, $p$ is
 the order of the p-norm (for $p \rightarrow \infty$, it becomes the commonly used max pooling). If
 $r > k$, pooling regions do not overlap; however, current CNN architectures typically include overlapping pooling with $k = 3$ and $r = 2$.
 Let us now compare the pooling operation defined by
 Eq. \ref{eq:pooling} to the standard definition of a
 convolutional layer $c$ applied to feature map $f$ given as:
 \begin{equation}
   c_{i,j,o}(f) = \sigma \left ( \sum_{h=-\floor{k/2}}^{\floor{k/2}}
     \sum_{w=-\floor{k/2}}^{\floor{k/2}} \sum_{u=1}^{N}
     \theta_{h,w,u,o} \cdot f_{g(h,w,i,j,u)} \right ),
  \label{eq:convolition}
 \end{equation}
 where $\mathbf{\theta}$ are the convolutional weights (or the kernel weights, or filters),
 $\sigma(\cdot)$ is the activation function, typically a rectified linear activation ReLU $\sigma(x) =
 \max(x, 0)$, and $o \in [1,M]$ is the number of output feature (or channel) of the
 convolutional layer. When
 formalized like this it becomes clear that both operations depend on the same elements of the 
 previous layer feature map. The pooling layer can be seen as performing a feature-wise convolution
\footnote{That is, a convolution where $\theta_{h,w,u,o} = 1$ if $u$
  equals $o$ and zero otherwise.} 
in which the
 activation function is replaced by the p-norm. 
 One can therefore ask the question whether and why such special layers need to be introduced into the network.
 While a complete answer of this question is not easy to give (see the
 experiments and discussion for further details and remarks) we assume that in general there exist three possible explanations why
%We see three possible explanations of why 
pooling can help in CNNs: 1) the p-norm makes the representation in a CNN more invariant; 2) the spatial dimensionality
reduction performed by pooling makes covering larger 
parts of the input in higher layers possible; 3) the feature-wise
nature of the pooling operation (as opposed to a convolutional layer where features get mixed) could make optimization easier.
Assuming that only the second part -- the dimensionality reduction
performed by pooling -- is crucial for achieving good performance with
CNNs (a hypothesis that we later test in our experiments) one can now
easily see that pooling can be removed from a network without
abandoning the spatial dimensionality reduction by two means:
\begin{enumerate}
  \item We can remove each pooling layer and increase the stride of the
convolutional layer that preceded it accordingly.
  \item We can replace the pooling layer by a normal convolution with
    stride larger than one (i.e. for a pooling layer with $k=3$ and
    $r=2$ we replace it with a convolution layer with corresponding stride
    and kernel size and number of output channels equal to the number of input channels)
    %and chose $N$ to be equivalent to $M$) .
\end{enumerate}
The first option has the downside that we significantly reduce the
overlap of the convolutional layer that preceded the pooling layer. It
is equivalent to a pooling operation in which only the top-left
feature response is considered and can result in less accurate
recognition. The second option does not suffer from this problem,
since all existing convolutional layers stay unchanged, but results in
an increase of overall network parameters. It is worth noting that
replacing pooling by convolution adds inter-feature dependencies
unless the weight matrix 
$\mathbf{\theta}$
is constrained.
We emphasize that that this
replacement can also be seen as 
learning the pooling operation rather than fixing it; which has
previously been considered using different parameterizations in the
literature \footnote{Although in order
  to implement ``proper pooling'' in the same sense as commonly
  considered in the literature a special nonlinearity (e.g. a squaring operation) needs to be
  considered. A simple convolution layer with rectified linear
  activation cannot by itself implement a p-norm computation.}
\citep{LeCun_IEEE1998,Gulcehre_2014,Jia_2012}.
We will evaluate both options in our experiments, ensuring a fair
comparison w.r.t. the number of network parameters. 
Although we are not aware of existing studies   
containing such controlled experiments on replacing pooling with convolution layers
it should be noted that the idea of removing pooling is not entirely unprecedented: 
First, the nomenclature in early work on CNNs \cite{LeCun_IEEE1998} (referring to pooling layers  as 
subsampling layers already) suggests the usage of different operations for subsampling. 
Second, albeit only considering small networks, 
experiments on using only convolution layers (with occasional subsampling) in an architecture similar
to traditional CNNs already appeared in work on the ``neural abstraction pyramid''\cite{Behnke2003}.

The second difference of the network model we consider to standard
CNNs is that -- similar to models recently used for achieving
state-of-the-art performance in the ILSVRC-2012 competition~\citep{VGG_2014,GoingDeep_2014} -- 
we make use of small convolutional layers with $k < 5$ which can
greatly reduce the number of parameters in a network and thus serve as
a form of regularization. 
Additionally, to unify the architecture further, we make use of the
fact that if the image area covered by units in the topmost
convolutional layer covers a portion of the image large enough to
recognize its content (i.e. the object we want to recognize) then
fully connected layers can also be replaced by simple 1-by-1
convolutions. This leads to predictions of object classes at different
positions which can then simply be averaged over the whole image. This
scheme was first described by \citet{Lin_2014} and
further regularizes the network as the one by one convolution has much
less parameters than a fully connected layer.
Overall our architecture is thus reduced to consist only of
convolutional layers with rectified linear non-linearities and an
averaging + softmax layer to produce predictions over the whole image.
%In-line with the general goal of
%finding a minimal architecture for recongition we consider the
%smallest possible 

%\subsection{Weight initialization}
%Blabla describe what exactly we ended up doing...

\begin{table}[h]
\caption{The three base networks used for classification on CIFAR-10
  and CIFAR-100.}
\label{base-models}
\begin{center}
\begin{small}
\begin{tabular}{l|l|l}
\multicolumn{3}{c}{\bf Model} \\
\hline
A         &B & C \\
\hline
\multicolumn{3}{c}{Input $32 \times 32$ RGB image} \\
\hline
$5 \times 5$ conv. $96$ ReLU & $5 \times 5$ conv. $96$ ReLU & $3 \times 3$ conv. $96$ ReLU \\
 & $1 \times 1$ conv. $96$ ReLU & $3 \times 3$ conv. $96$ ReLU \\
\hline 
\multicolumn{3}{c}{$3 \times 3$ max-pooling stride $2$} \\ 
\hline
$5 \times 5$ conv. $192$ ReLU & $5 \times 5$ conv. $192$ ReLU & $3 \times 3$ conv. $192$ ReLU \\
 & $1 \times 1$ conv. $192$ ReLU & $3 \times 3$ conv. $192$ ReLU \\
\hline
\multicolumn{3}{c}{$3 \times 3$ max-pooling stride $2$} \\
\hline
\multicolumn{3}{c}{$3 \times 3$ conv. $192$ ReLU} \\
\hline
\multicolumn{3}{c}{$1 \times 1$ conv. $192$ ReLU} \\
\hline
\multicolumn{3}{c}{$1 \times 1$ conv. $10$ ReLU} \\
\hline
\multicolumn{3}{c}{global averaging over $6\times6$ spatial dimensions} \\
\hline
\multicolumn{3}{c}{10 or 100-way softmax} \\
\end{tabular}
\end{small}
\end{center}
\end{table}

\section{Experiments}
In order to quantify the effect of simplifying the model architecture
we perform experiments on three datasets: CIFAR-10, CIFAR-100~\citep{Krizhevsky2009} and ILSVRC-2012 ImageNet~\citep{Imagenet_2009}\,. Specifically, we use CIFAR-10 to perform an in-depth
study of different models, since a large model on this dataset can be trained with
moderate computing costs of $\approx 10$ hours on a modern GPU. We
then test the best model found on CIFAR-10 and CIFAR-100 with and
without augmentations and perform a first preliminary experiment on
the ILSVRC-2012 ImageNet dataset. We performed all experiments using the \emph{Caffe}~\citep{caffe} framework.

\subsection{Experimental Setup}
\label{sect:setup}
In experiments on CIFAR-10 and CIFAR-100 we use three different base network models which are intended to
reflect current best practices for setting up CNNs for object
recognition. 
Architectures of these networks are described in Table
\ref{base-models}. Starting from model A (the simplest model) the depth and
number of parameters in the network gradually increases to model C. 
Several things are to be noted here. First, as described in the table,
all base networks we consider use a 1-by-1 convolution at the top to
produce 10 outputs of which we then compute an average over all
positions and a softmax to produce class-probabilities (see Section
\ref{sect:model} for the rationale behind this approach). We performed
additional experiments with fully connected layers instead of 
1-by-1 convolutions but found these models to consistently perform
$ 0.5 \% - 1 \%$ worse than their fully convolutional counterparts. This
is in line with similar findings from prior work~\citep{Lin_2014}. We
hence do not report these numbers here to avoid cluttering the
experiments. Second, it can be observed that model B from the table is
a variant of the Network in Network architecture proposed by \citet{Lin_2014} in which only one 1-by-1 convolution is
performed after each ``normal'' convolution layer. Third, model C
replaces all $5 \times 5$ convolutions by simple $3 \times 3$
convolutions. This serves two purposes: 1) it unifies the
architecture to consist only of layers operating on $3 \times 3$ spatial
neighborhoods of the previous layer feature map (with occasional
subsampling); 2) if max-pooling is replaced by a
convolutional layer, then $3 \times 3$ is the minimum filter size to allow overlapping convolution with stride 2. 
%(see Section \ref{sect:model} for a more thorough discussion). 
We also highlight that model C resembles the very deep
models used by \citet{VGG_2014} in this years ImageNet
competition.

\begin{table}[h]
\caption{Model description of the three networks derived from base model C used for evaluating the importance of pooling in case of classification on CIFAR-10 and CIFAR-100. The derived models for base models A and B are built analogously. The higher layers are the same as in Table \ref{base-models}~.}
\label{derived-models}
\begin{center}
\begin{small}
\begin{tabular}{l|l|l}
\multicolumn{3}{c}{\bf Model} \\
\hline
Strided-CNN-C         &  ConvPool-CNN-C & All-CNN-C \\
\hline
\multicolumn{3}{c}{Input $32 \times 32$ RGB image} \\
\hline
$3 \times 3$ conv. $96$ ReLU  & $3 \times 3$ conv. $96$ ReLU  & $3 \times 3$ conv. $96$ ReLU  \\
$3 \times 3$ conv. $96$ ReLU  & $3 \times 3$ conv. $96$ ReLU  & $3 \times 3$ conv. $96$ ReLU  \\
\multicolumn{1}{c|}{with stride $r = 2$}    & $3 \times 3$ conv. $96$ ReLU  &  \\
\hline 
                                   & $3 \times 3$ max-pooling stride $2$ & $3 \times 3$ conv. $96$ ReLU  \\
                                   &                                     & \multicolumn{1}{c}{with stride $r = 2$} \\
\hline
$3 \times 3$ conv. $192$ ReLU & $3 \times 3$ conv. $192$ ReLU & $3 \times 3$ conv. $192$ ReLU \\
$3 \times 3$ conv. $192$ ReLU & $3 \times 3$ conv. $192$ ReLU  & $3 \times 3$ conv. $192$ ReLU \\
\multicolumn{1}{c|}{with stride $r = 2$}    & $3 \times 3$ conv. $192$ ReLU &  \\
\hline
                                   & $3 \times 3$ max-pooling stride $2$ & $3 \times 3$ conv. $192$ ReLU \\
                                   &                                     & \multicolumn{1}{c}{with stride $r = 2$} \\
\hline
\multicolumn{3}{c}{$\vdots$} \\
%\hline
%\multicolumn{3}{c}{$1 \times 1$ conv $192$ ReLU} \\
%\hline
%\multicolumn{3}{c}{$1 \times 1$ conv $10$ ReLU} \\
%\hline
%\multicolumn{3}{c}{global averaging over $6\times6$ spatial dimensions} \\
%\hline
%\multicolumn{3}{c}{10 or 100-way softmax} \\
\end{tabular}
\end{small}
\end{center}
\end{table}

For each of the base models we then experiment with three additional variants. The additional (derived) models for base model C are described in in Table \ref{derived-models}. The derived models for base models A and B are built analogously but not shown in the table to avoid cluttering the paper. 
%(please see the appendix for a complete table of all models). 
In general the additional models for each base model consist of:
\begin{itemize}
  \item A model in which max-pooling is removed and the stride of the convolution layers
preceding the max-pool layers is increased by 1 (to ensure that the
next layer covers the same spatial region of the input image as
before). This is column ``Strided-CNN-C'' in the table.  
  \item A model in which max-pooling is replaced by a convolution
    layer. This is column ``All-CNN-C'' in the table.
  \item A model in which a dense convolution is placed before each
    max-pooling layer (the additional convolutions have the same kernel size as
    the respective pooling layer). This is model ``ConvPool-CNN-C'' in
    the table. Experiments with this model are necessary to ensure
    that the effect we measure is not solely due to increasing model
    size when going from a ``normal'' CNN to its ``All-CNN'' counterpart.
\end{itemize}

Finally, to test whether a network solely using convolutions also
performs well on a larger scale recognition problem we trained an
up-scaled version of ALL-CNN-B on the ILSVRC 2012 part of the ImageNet
database. Although we
expect that a larger network using only $3 \times 3$ convolutions and having stride $1$ in the first layer (and 
thus similar in style to~\citet{VGG_2014}) would perform even
better on this dataset, training it would take several weeks 
and could thus not be completed in time for this manuscript.

\subsection{Classification results}
\subsubsection{CIFAR-10}
\begin{table}[h]
\caption{Comparison between the base and derived models on the CIFAR-10 dataset.}
\label{results-cifar10-base}
\begin{center}
\begin{tabular}{lll}
\multicolumn{3}{c}{\bf CIFAR-10 classification error} \\
\multicolumn{1}{l}{Model} & \multicolumn{1}{l}{Error ($\%$)} & \multicolumn{1}{l}{\# parameters} \\
\hline
\multicolumn{3}{l}{without data augmentation} \\
\hline
Model A         &  $12.47 \%$ & $\approx 0.9$ M \\
Strided-CNN-A         &  $13.46 \%$ & $\approx 0.9$ M \\
ConvPool-CNN-A         &  $\mathbf{10.21} \%$ & $\approx 1.28$ M \\
ALL-CNN-A         &  $10.30 \%$ & $\approx 1.28$ M \\
\hline
Model B         &  $10.20 \%$ & $\approx 1$ M \\
Strided-CNN-B         &  $10.98 \%$ & $\approx 1$ M \\
ConvPool-CNN-B         &  $9.33 \%$ & $\approx 1.35$ M \\
ALL-CNN-B         &  $\mathbf{9.10} \%$ & $\approx 1.35$ M \\
\hline
Model C         &  $9.74 \%$ & $\approx 1.3$ M \\
Strided-CNN-C         &  $10.19 \%$ & $\approx 1.3$ M \\
ConvPool-CNN-C         &  $9.31 \%$ & $\approx 1.4$ M \\
ALL-CNN-C         &  $\mathbf{9.08 \%}$ & $\approx 1.4$ M \\
\end{tabular}
\end{center}
\end{table}

In our first experiment we compared all models from Section
\ref{sect:setup} on the CIFAR-10 dataset without using any
augmentations. All networks were trained using stochastic gradient
descent with fixed momentum of $0.9$. The learning rate $\gamma$ was adapted
using a schedule $S = {e_1, e_2, e_3}$ in which $\gamma$ is multiplied by a fixed multiplier
of $0.1$ after $e_1. e_2$ and $e_3$ epochs respectively. 
To keep the amount of computation necessary to perform
our comparison bearable~\footnote{Training one network on CIFAR-10 can
  take up to 10 hours on a modern GPU.} we only treat $\gamma$ as a
changeable hyperparameter for each method. The learning rate schedule
and the total amount of training epochs were determined in a preliminary
experiment using base model A and then fixed for all other
experiments. We used $S = [200, 250, 300]$ and trained all networks
for a total of 350 epochs. It should be noted that this strategy is not guaranteed to result in the best
performance for all methods and thus care must be taken when
interpreting the the following results from our experiments.
The learning rate $\gamma$ was individually adapted for each model by
searching over the fixed set $\gamma \in [0.25, 0.1, 0.05,
0.01]$. In the following we only report the results for the best
$\gamma$ for each method. 
Dropout~\citep{Hinton_arxiv2012} was used to regularize all networks. We
applied dropout to the input image as well as after each pooling layer
(or after the layer replacing the pooling layer respectively). The
dropout probabilities were $20 \%$ for dropping out inputs and $50 \%$
otherwise. We also experimented with additional dropout (i.e. dropout
on all layers or only on the $1\times1$ convolution layer) which
however did not result in increased accuracy\footnote{In the case were
dropout of 0.5 is applied to all layers accuracy even dropped,
suggesting that the gradients become too noisy in this case}~. 
Additionally all models were regularized with weight decay $\lambda =
0.001$. In experiments with data augmentation we perform only the
augmentations also used in previous work
\citep{Goodfellow2013,Lin_2014} in order to keep our results
comparable. These include adding horizontally flipped examples of
all images as well as randomly translated versions (with a maximum
translation of 5 pixels in each dimension). In all experiments images
were whitened and contrast normalized following \citet{Goodfellow2013}.

The results for all models that we considered are given in Table \ref{results-cifar10-base}.
Several trends can be observed from the table. First, confirming
previous results from the literature \citep{Srivastava14a} the simplest
model (model A) already performs remarkably well, achieving $12.5 \%$
error. Second, simply removing the max-pooling layer and just increasing the
stride of the previous layer results in diminished performance in all
settings. While this is to be expected we can already see that the
drop in performance is not as dramatic as one might expect from such a
drastic change to the network architecture.
Third, surprisingly, when pooling is replaced by an additional convolution layer
with stride $r = 2$ performance stabilizes and even improves on the
base model. To check that this is not only due to
an increase in the number of trainable parameters we compare the results to the \mbox{``ConvPool''}
versions of the respective base model. In all cases the performance of
the model without any pooling and the model with pooling on top of the
additional convolution perform about on par. Surprisingly, this suggests that while
pooling can help to regularize CNNs, and generally does not hurt
performance, it is not strictly necessary to achieve state-of-the-art
results (at least for current small scale object recognition
datasets). In addition, our results confirm that small $3\times3$
convolutions stacked after each other seem to be enough to achieve the
best performance.

Perhaps even more interesting is the comparison between the simple all
convolutional network derived from base model C and the state of the art on CIFAR-10 shown in Table
\ref{results-cifar10}~, both with and without data augmentation. In both
cases the simple network performs better than the best previously
reported result.
This suggests that in order to perform well on
current benchmarks ``almost all you need'' is a stack of convolutional
layers with occasional stride of $2$ to perform subsampling. 

\begin{table}[h]
\caption{Test error on CIFAR-10 and CIFAR-100 for the All-CNN compared to the
  state of the art from the literature. The All-CNN is the version
  adapted from base model C (i.e. All-CNN-C). The other results are from: [1]
  \citep{Goodfellow2013}, [2] \citep{Lin_2014}, [3] \citep{Lee_2014},
  [4] \citep{Stollenga_2014}, [5] \citep{Nitish2013}, [6] \citep{Graham2015}. The number of
  parameters is given in million parameters.}
\label{results-cifar10}
%\begin{center}
\begin{minipage}{0.49\linewidth}
\centering
\begin{tabular}{lll}
\multicolumn{3}{c}{\bf CIFAR-10 classification error} \\
\multicolumn{1}{l}{Method} & \multicolumn{1}{l}{Error ($\%$)} & \multicolumn{1}{l}{\# params} \\
\hline
\multicolumn{3}{l}{without data augmentation} \\
\hline
Maxout [1]          &  $11.68 \%$ & $> 6$ M \\
Network in Network [2]         &  $10.41 \%$ & $\approx 1$ M \\
Deeply Supervised [3]        & $9.69 \%$  & $\approx 1$ M \\
\textbf{ALL-CNN (Ours)}         &  $\mathbf{9.08 \%}$ & $\approx 1.3$ M \\
\hline
\multicolumn{3}{l}{with data augmentation} \\
\hline
Maxout [1]         &  $9.38 \%$ & $> 6$ M \\
DropConnect [2]         &  $9.32 \%$ & - \\
dasNet [4]         &  $9.22 \%$ & $> 6$ M \\
Network in Network [2]         &  $8.81 \%$ & $\approx 1$ M \\
Deeply Supervised [3]         & $7.97 \%$  & $\approx 1$ M \\
\textbf{ALL-CNN (Ours)}         &  $\mathbf{7.25 \%}$ & $\approx 1.3$ M \\
\end{tabular}
\end{minipage}
\hspace{0.5cm}
\begin{minipage}{0.49\linewidth}
\centering
\begin{tabular}{ll}
\multicolumn{2}{c}{\bf CIFAR-100 classification error } \\
Method & Error ($\%$) \\ %& \# params \\
\hline
%Maxout [1]         &  $38.57 \%$ \\ %& $\approx 6$ M \\
CNN + tree prior [5]         &  $36.85 \%$ \\ %& $> 10$ M \\
Network in Network [2]         &  $35.68 \%$ \\% & $\approx 1$ M \\
Deeply Supervised [3]         & $34.57 \%$  \\ % & $\approx 1$ M \\
Maxout (larger) [4]         &  $34.54 \%$ \\ % & $> 10$ M \\
dasNet [4]         &  $33.78 \%$ \\% & $> 10$ M \\
ALL-CNN (Ours)         &  $33.71 \%$ \\% & $\approx 1.3$ M \\
Fractional Pooling (1 test) [6]  & $\mathbf{31.45} \%$ \\
\textbf{Fractional Pooling (12 tests) [6]}  & $\mathbf{26.39} \%$ \\
\end{tabular}
\begin{tabular}{ll}
\multicolumn{2}{c}{\bf CIFAR-10 classification error} \\
\multicolumn{1}{l}{Method} & \multicolumn{1}{l}{Error ($\%$)} \\
\hline
\multicolumn{2}{l}{with large data augmentation} \\
\hline
Spatially Sparse CNN [6]         & $4.47 \%$  \\
Large ALL-CNN (Ours)         &  $4.41 \%$  \\
Fractional Pooling (1 test) [6]  & $4.50 \%$ \\
\textbf{Fractional Pooling (100 tests) [6]}  & $\mathbf{3.47} \%$ \\
\end{tabular}
\end{minipage}
%\end{center}
\end{table}

%\begin{table}[h]
%\caption{Test error on CIFAR-100 for the All-CNN compared to the
%  state of the art from the literature. The All-CNN is the version adapted from base model C (i.e. All-CNN-C).}
%\label{results-cifar100}
%\begin{center}
%\begin{tabular}{l|l|l}
%\multicolumn{3}{c}{\bf CIFAR-100 classification error } \\
%\hline
%Method & Error ($\%$) & \# parameters \\
%\hline
%Maxout \citep{Goodfellow2013}         &  $38.57 \%$ & $\approx 6$ M \\
%Maxout (larger) \citep{Stollenga_2014}         &  $38.57 \%$ & $> 10$ M \\
%CNN + tree prior \citep{Nitish2013}         &  $36.85 \%$ & $> 10$ M \\
%Network in Network \citep{Lin_2014}         &  $35.68 \%$ & $\approx 1$ M \\
%Deeply Supervised \citep{Lee_2014}         & $34.57 \%$  & $\approx 1$ M \\
%dasNet \citep{Stollenga_2014}         &  $\mathbf{33.78 \%}$ & $> 10$ M \\
%\textbf{ALL-CNN (Ours)}         &  $\mathbf{33.71 \%}$ & $\approx 1.3$ M \\
%\end{tabular}
%\end{center}
%\end{table}
 
\subsubsection{CIFAR-100}
We performed an additional experiment on the CIFAR-100 dataset to
confirm the efficacy of the best model (the All-CNN-C) found for
CIFAR-10. As is common practice we used the same model as on CIFAR-10
and also kept all hyperparameters (the learning rate as well as its
schedule) fixed. Again note that this does not necessarily give the
best performance. The results of this experiment are given in Table
\ref{results-cifar10} (right). As can be seen, the simple model using only $3\times3$
convolutions again performs comparable to the state of the art
for this dataset even though most of the other methods either use more
complicated training schemes or network architectures. 
It is only outperformed by the fractional max-pooling approach \citep{Graham2015} which uses
a much larger network (on the order of $50 M$ parameters).

\subsubsection{CIFAR-10 with additional data augmentation}
After performing our experiments we became aware of recent results by \citet{Graham2015} who report
a new state of the art on CIFAR-10/100 with data augmentation. These results 
were achieved using very deep CNNs with $2 \times 2$ convolution layers in combination with aggressive data augmentation 
in which the $32 \times 32$ images are placed into large $126 \times 126$ pixel images and can hence be 
heavily scaled, rotated and color augmented. We thus implemented the Large-All-CNN, which is the all convolutional version of this network (see Table \ref{tbl:large_cifar_net} in the appendix for details) and 
report the results of this additional experiment in Table \ref{results-cifar10} (bottom right).
As can be seen, Large-All-CNN achieves performance comparable to the network with max-pooling. 
It is only outperformed by the fractional max-pooling approach when performing multiple passes through the network. Note
that these networks have vastly more parameters ($> 50$ M) than the
networks from our previous experiments.
We are currently re-training the Large-All-CNN network on CIFAR-100,
and will include the results in Table \ref{results-cifar10} once training is finished. 

\subsection{Classification of Imagenet}
We performed additional experiments using the ILVRC-2012 subset of the ImageNet dataset. Since training a state of the art model on this dataset can take several weeks of computation on a modern GPU, we did not aim for best performance, but rather performed a simple 'proof of concept' experiment. To test if the architectures performing best on CIFAR-10 also apply to larger datasets, we trained an upscaled version of the All-CNN-B network (which is also similar to the architecture proposed by~\citet{Lin_2014}). %The resulting network performs about as well as the network from~\citet{Krizhevsky_NIPS2012}, while training time is faster. 
It has 12 convolutional layers (conv1-conv12) and was trained for $450,000$ iterations with batches of $64$ samples each, starting with a learning rate of $\gamma = 0.01$ and dividing it by $10$ after every $200,000$ iterations. A weight decay of $\lambda = 0.0005$ was used in all layers. The exact architecture used is given in Table~\ref{tbl:imagenet_net} in the Appendix.

This network achieves a Top-1 validation error of $41.2 \%$ on ILSVRC-2012, when only evaluating on the center $224\times224$ patch,~--~which is comparable to the $40.7 \%$ Top-1 error reported by~\citet{Krizhevsky_NIPS2012}~--~while having less than $10$ million parameters (6 times less than the network of~\citet{Krizhevsky_NIPS2012}) and taking roughly $4$ days to train on a Titan GPU. This supports our intuition that max-pooling may not be necessary for training large-scale convolutional networks. However, a more thorough analysis is needed to precisely evaluate the effect of max-pooling on ImageNet-scale networks. Such a complete quantitative analysis using multiple networks on ImageNet is extremely computation-time intensive and thus out of the scope of this paper. In order to still gain some insight into the effects of getting rid of max-pooling layers, we will try to analyze the representation learned by the all convolutional network in the next section.

\subsection{Deconvolution}
In order to analyze the network that we trained on ImageNet -- and get a first impression of how well the model without pooling lends itself to approximate inversion -- we use a 'deconvolution' approach. We start from the idea of using a deconvolutional network for visualizing the parts of an image that are most discriminative for a given unit in a network, an approach recently proposed by \citet{Zeiler_ECCV2014}. Following this initial attempt -- and observing that it does not always work well without max-pooling layers -- we propose a new and efficient way of visualizing the concepts learned by higher network layers. 

\begin{figure}[h!]
\begin{center}
\includegraphics[width=.7\textwidth]{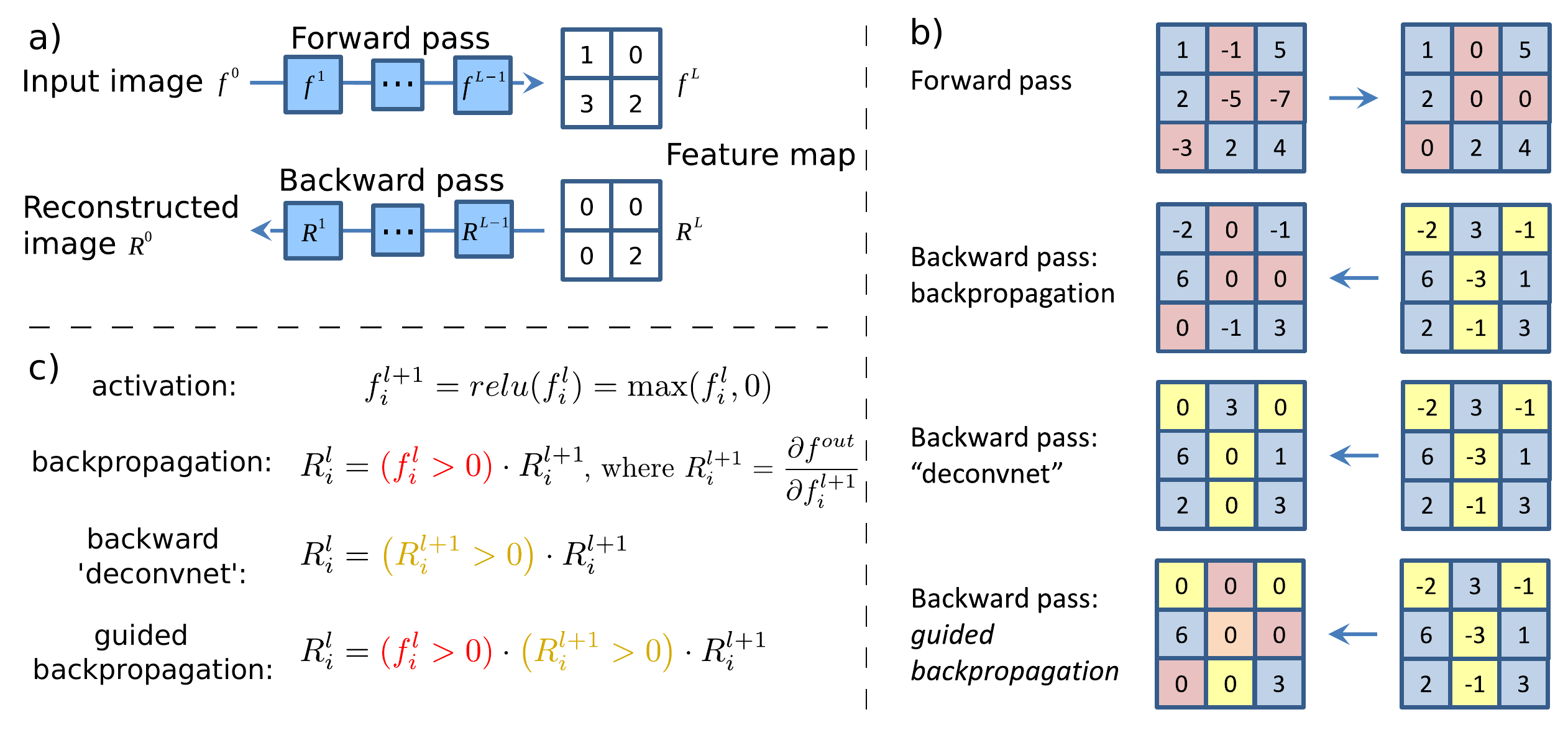}
\end{center}
\caption{Schematic of visualizing the activations of high layer
  neurons. a) Given an input image, we perform the forward pass to the
  layer we are interested in, then set to zero all activations except
  one and propagate back to the image to get a reconstruction. b)
  Different methods of propagating back through a ReLU
  nonlinearity. c) Formal definition of different methods for
  propagating a output activation $out$ back through a ReLU unit in
  layer $l$; note that the 'deconvnet' approach and guided backpropagation 
  do not compute a true gradient but rather an imputed version.}
\label{fig:scheme_deconv}
\end{figure}

\begin{figure}
\hspace*{-.0cm}
\begin{center}
\begin{tabular}{>{\centering\arraybackslash} m{3.3cm} >{\centering\arraybackslash} m{3.3cm} >{\centering\arraybackslash} m{4.3cm}}
  conv1 & conv2 & conv3 \\
  \includegraphics[width=.23\textwidth]{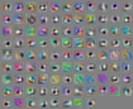} &
  \includegraphics[width=.23\textwidth]{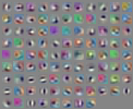} &
  \includegraphics[width=.3\textwidth]{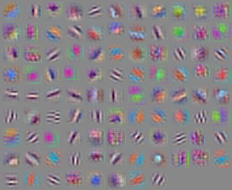} \\
\end{tabular}
\end{center}
\vspace*{-0.2cm}
\caption{Visualizations of patterns learned by the lower layers (conv1-conv3) of the network trained on ImageNet. Each single patch corresponds to one filter. Interestingly, Gabor filters only appear in the third layer.}
\label{fig:lower_layer_reconstructions}
\vspace*{-0.2cm}
\end{figure}

The deconvolutional network ('deconvnet') approach to visualizing concepts learned by neurons in higher layers of a CNN can be summarized as follows. Given a high-level feature map, the 'deconvnet' inverts the data flow of a CNN, going from neuron activations in the given layer down to an image. Typically, a single neuron is left non-zero in the high level feature map. Then the resulting reconstructed image shows the part of the input image that is most strongly activating this neuron (and hence the part that is most discriminative to it). A schematic illustration of this procedure is shown in Figure~\ref{fig:scheme_deconv}~a).
In order to perform the reconstruction through max-pooling layers, which are in general not invertible, the method of Zeiler and Fergus requires first to perform a forward pass of the network to compute 'switches' -- positions of maxima within each pooling region. These switches are then used in the 'deconvnet' to obtain a discriminative reconstruction. By using the switches from a forward pass the 'deconvnet' (and thereby its reconstruction) is hence conditioned on an image and does \emph{not} directly visualize learned features. Our architecture does not include max-pooling, meaning that in theory we can 'deconvolve' without switches, i.e. not conditioning on an input image. 
This way we get insight into what lower layers of the network learn. Visualizations of features from the first three layers are shown in Figure~\ref{fig:lower_layer_reconstructions}~. Interestingly, the very first layer of the network does not learn the usual Gabor filters, but higher layers do.

For higher layers of our network the method of Zeiler and Fergus fails to produce sharp, recognizable, image structure. This is in agreement with the fact that lower layers learn general features with limited amount of invariance, which allows to reconstruct a single pattern that activates them. However, higher layers learn more invariant representations, and there is no single image maximally activating those neurons. Hence to get reasonable reconstructions it is necessary to condition on an input image.

%\begin{figure}
%\begin{center}
%\includegraphics[width=.3\textwidth]{scheme_relu_methods.pdf}
%\end{center}
%\caption{Different methods of propagating back through a ReLU nonlinearity.}
%\label{fig:scheme_relu_methods}
%\end{figure}

An alternative way of visualizing the part of an image that most activates a given neuron is to use a simple backward pass of the activation of a single neuron after a forward pass through the network; thus computing the gradient of the activation w.r.t. the image. The backward pass is, by design, partially conditioned on an image through both the activation functions of the network and the max-pooling switches (if present). The connections between the deconvolution and the backpropagation approach were recently discussed in \citet{Simonyan_arxiv2014}. In short the both methods differ mainly in the way they handle backpropagation through the rectified linear (ReLU) nonlinearity.

\begin{figure}
\hspace*{-.0cm}
\begin{tabular}{ccc}
  deconv & guided backpropagation & corresponding image crops \\
  \includegraphics[height=.25\textwidth]{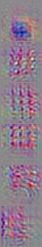} &
  \includegraphics[height=.25\textwidth]{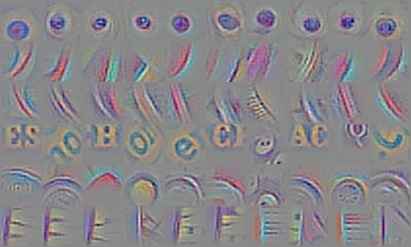} &
  \includegraphics[height=.25\textwidth]{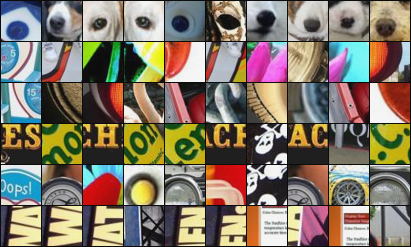} \\
\end{tabular}
\hspace*{-.0cm}
\begin{tabular}{ccc}
  deconv & guided backpropagation & corresponding image crops \\
  \includegraphics[height=.25\textwidth]{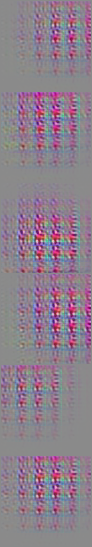} &
  \includegraphics[height=.25\textwidth]{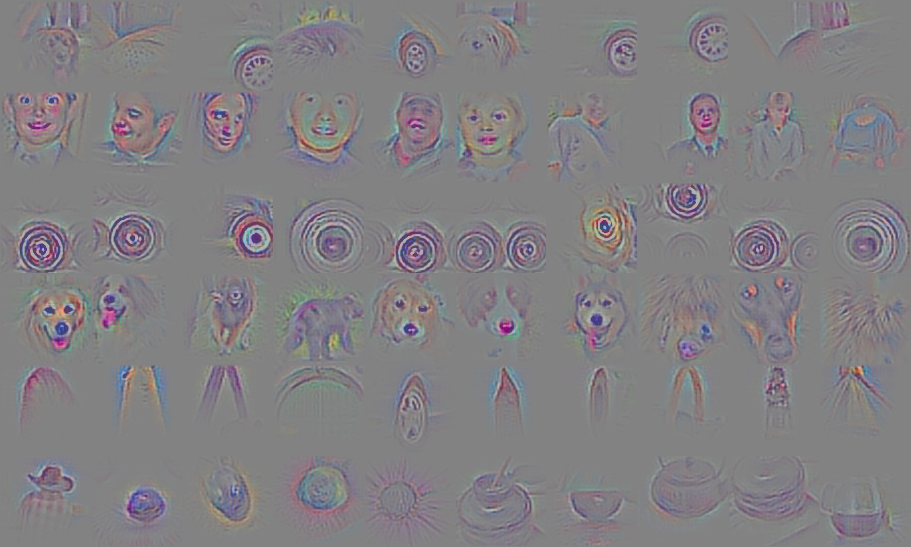} &
  \includegraphics[height=.25\textwidth]{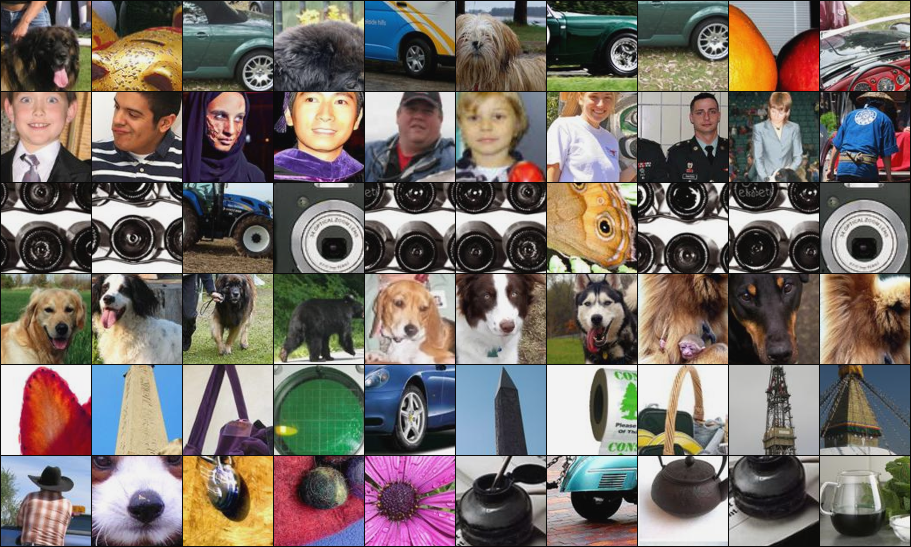} \\
\end{tabular}
\caption{Visualization of patterns learned by the layer conv6 (top)
  and layer conv9 (bottom) of the network trained on ImageNet. Each
  row corresponds to one filter. The
  visualization using ``guided backpropagation'' is based on the top 10 image
  patches activating this filter taken from the ImageNet dataset. Note
  that image sizes are not preserved (in order to save space).}
\label{fig:pool_reconstructions}
\end{figure}

In order to obtain a reconstruction conditioned on an input image from our network without pooling layers we propose a modification of the 'deconvnet', which makes reconstructions significantly more accurate, especially when reconstructing from higher layers of the network. The 'deconvolution' is equivalent to a backward pass through the network, except that when propagating through a nonlinearity, its gradient is solely computed based on the top gradient signal, ignoring the bottom input. In case of the ReLU nonlinearity this amounts to setting to zero certain entries based on the top gradient. The two different approaches are depicted in Figure~\ref{fig:scheme_deconv}~b), rows 2 and 3. We propose to combine these two methods: rather than masking out values corresponding to negative entries of the top gradient ('deconvnet') or bottom data (backpropagation), we mask out the values for which at least one of these values is negative, see row 4 of Figure~\ref{fig:scheme_deconv}~b). We call this method \emph{guided backpropagation}, because it adds an additional guidance signal from the higher layers to usual backpropagation. This prevents backward flow of negative gradients, corresponding to the neurons which decrease the activation of the higher layer unit we aim to visualize. Interestingly, unlike the 'deconvnet', guided backpropagation works remarkably well without switches, and hence allows us to visualize intermediate layers (Figure~\ref{fig:pool_reconstructions}) as well as the last layers of our network (Figures~\ref{fig:cccp8_different_methods} and~\ref{fig:pool4_different_methods} in the Appendix). In a sense, the bottom-up signal in form of the pattern of bottom ReLU activations substitutes the switches. 

To compare guided backpropagation and the 'deconvnet' approach, we replace the stride in our network by $2\times 2$ max-pooling \emph{after training}, which allows us to obtain the values of switches. We then visualize high level activations using three methods: backpropagation, 'deconvnet' and guided backpropagation. A striking difference in image quality is visible in the feature visualizations of the highest layers of the network, see Figures~\ref{fig:cccp8_different_methods} and~\ref{fig:pool4_different_methods} in the Appendix. Guided backpropagation works equally well with and without switches, while the 'deconvnet' approach fails completely in the absence of switches. One potential reason why the 'deconvnet' underperforms in this experiment is that max-pooling was only 'artificially' introduced after training. As a control Figure~\ref{fig:fc8_caffenet} shows visualizations of units in the fully connected layer of a network initially trained with max-pooling. Again guided backpropagation produces cleaner visualizations than the 'deconvnet' approach.

\section{Discussion}
To conclude, we highlight a few key observations that we made
in our experiments:
\begin{itemize}
 \item With modern methods of training convolutional neural networks very simple architectures may perform very well: a network using nothing but convolutions and subsampling matches or even slightly outperforms the state of the art on CIFAR-10 and CIFAR-100. A similar architecture shows competitive results on ImageNet.
 \item In particular, as opposed to previous observations, including explicit (max-)pooling operations in a network does not always improve performance of CNNs. This seems to be especially the case if the network is large enough for the dataset it is being trained on and can learn all necessary invariances just with convolutional layers.
 \item We propose a new method of visualizing the representations learned by higher layers of a convolutional network. While being very simple, it produces sharper visualizations of descriptive image regions than the previously known methods, and can be used even in the absence of 'switches'~-- positions of maxima in max-pooling regions.
\end{itemize}
We want to emphasize that this paper is not meant to discourage the use of pooling or more sophisticated activation functions altogether. It should rather be understood as an attempt to both search for the minimum necessary ingredients for  recognition with CNNs and establish a strong baseline on often used datasets. 
%TOBI: I added this part above and the part below. The one above I would definitely like to have, the part below is a bit hand wavey but somehting like this might be a nice finish to the paper. Any better ideas how to write this ?
We also want to stress that the results of all models evaluated in this paper could potentially be improved by increasing the overall model size or a more thorough hyperparameter search. In a sense this fact makes it even more surprising that the simple model outperforms many existing approaches. %and can be seen as a testament for the need of better tools to explore and compare the effectiveness of different network architectures more quickly. 

%\paragraph{Importance of different components in a CNN} We did not
%find any significant advantage in 

\section*{Acknowledgments}
We acknowledge funding by the ERC Starting Grant VideoLearn (279401); the work was also partly supported by the BrainLinks-BrainTools Cluster of
Excellence funded by the German Research Foundation (DFG, grant number EXC 1086).

%\newpage

\bibliography{iclr2015}
\bibliographystyle{iclr2015}

%\newpage

\section*{Appendix}
\begin{appendix}

\section{Large All-CNN Model for CIFAR-10}
The complete model architecture for the large All-CNN derived from the spatially sparse network of Benjamin Graham (see \citet{Graham2015} for an explanation) is givenin Table \ref{tbl:large_cifar_net}~. Note that the network uses leaky ReLU units instead of ReLUs as we found these to speed up training. As can be seen it also requires a much larger input size in which the $32 \times 32$ pixel image is centered (and then potentially augmented by applying multiple transformations such as scaling). As a result the subsampling performed by the convolutional layers with stride 2 can hence be applied much more slowly. Also note that this network only consists of $2 \times 2$ convolutions with occasional subsampling until the spatial dimensionality is reduced to $1\times 1$. It does hence not employ global average pooling at the end of the network. In a sense this architecture hence represents the most simple convolutional network usable for this task.
\begin{table}[h]
\caption{Architecture of the Large All-CNN network for CIFAR-10.}
\label{tbl:large_cifar_net}
\begin{center}
\begin{tabular}{c|c}
\multicolumn{2}{c}{\textbf{Large All-CNN for CIFAR-10}} \\ \hline
Layer name & Layer description \\ \hline
input & Input $126 \times 126$ RGB image\\
conv1 & $2 \times 2$ conv. 320 LeakyReLU, stride 1\\
conv2 & $2 \times 2$ conv. 320 LeakyReLU, stride 1\\
conv3 & $2 \times 2$ conv. 320 LeakyReLU, stride 2\\
conv4 & $2 \times 2$ conv. 640 LeakyReLU, stride 1, dropout $0.1$\\
conv5 & $2 \times 2$ conv. 640 LeakyReLU, stride 1, dropout $0.1$\\
conv6 & $2 \times 2$ conv. 640 LeakyReLU, stride 2 \\
conv7 & $2 \times 2$ conv. 960 LeakyReLU, stride 1, dropout $0.2$\\
conv8 & $2 \times 2$ conv. 960 LeakyReLU, stride 1, dropout $0.2$\\
conv9 & $2 \times 2$ conv. 960 LeakyReLU, stride 2 \\
conv10 & $2 \times 2$ conv. 1280 LeakyReLU, stride 1, dropout $0.3$\\
conv11 & $2 \times 2$ conv. 1280 LeakyReLU, stride 1, dropout $0.3$\\
conv12 & $2 \times 2$ conv. 1280 LeakyReLU, stride 2 \\
conv13 & $2 \times 2$ conv. 1600 LeakyReLU, stride 1, dropout $0.4$\\
conv14 & $2 \times 2$ conv. 1600 LeakyReLU, stride 1, dropout $0.4$\\
conv15 & $2 \times 2$ conv. 1600 LeakyReLU, stride 2 \\
conv16 & $2 \times 2$ conv. 1920 LeakyReLU, stride 1, dropout $0.5$\\
conv17 & $1 \times 1$ conv. 1920 LeakyReLU, stride 1, dropout $0.5$\\
softmax & 10-way softmax
\end{tabular}
\end{center}
\end{table}

\section{Imagenet Model}
The complete model architecture for the network trained on the ILSVRC-2102 ImageNet dataset is given in Table \ref{tbl:imagenet_net}~.
\newpage
\begin{table}[ht]
\caption{Architecture of the ImageNet network.}
\label{tbl:imagenet_net}
\begin{center}
\begin{tabular}{c|c}
\multicolumn{2}{c}{\textbf{ImageNet model}} \\ \hline
Layer name & Layer description \\ \hline
input & Input $224 \times 224$ RGB image\\
conv1 & $11 \times 11$ conv. 96 ReLU units, stride 4\\
conv2 & $1 \times 1$ conv. 96 ReLU, stride 1\\
conv3 & $3 \times 3$ conv. 96 ReLU, stride 2\\
conv4 & $5 \times 5$ conv. 256 ReLU, stride 1\\
conv5 & $1 \times 1$ conv. 256 ReLU, stride 1\\
conv6 & $3 \times 3$ conv. 256 ReLU, stride 2\\
conv7 & $3 \times 3$ conv. 384 ReLU, stride 1\\
conv8 & $1 \times 1$ conv. 384 ReLU, stride 1\\
conv9 & $3 \times 3$ conv. 384 ReLU, stride 2, dropout 50 \%\\
conv10 & $3 \times 3$ conv. 1024 ReLU, stride 1\\
conv11 & $1 \times 1$ conv. 1024 ReLU, stride 1\\
conv12 & $1 \times 1$ conv. 1000 ReLU, stride 1\\
global\_pool & global average pooling ($6 \times 6$)\\
softmax & 1000-way softmax
\end{tabular}
\end{center}
\end{table}

\section{Additional Visualizations}
\begin{figure}[h]
\begin{center}
\begin{tabular}{ >{\centering\arraybackslash} m{1.5cm} >{\centering\arraybackslash} m{3.5cm} >{\centering\arraybackslash} m{3.5cm} >{\centering\arraybackslash} m{3.5cm} }
   & backpropagation & 'deconvnet' & guided backpropagation \\
  with pooling + switches & 
  \includegraphics[width=.27\textwidth]{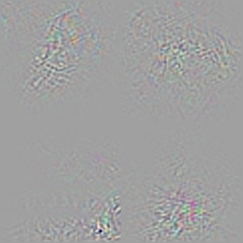} &
  \includegraphics[width=.27\textwidth]{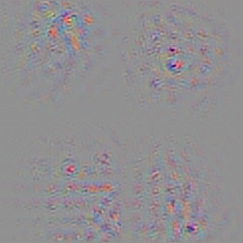} &
  \includegraphics[width=.27\textwidth]{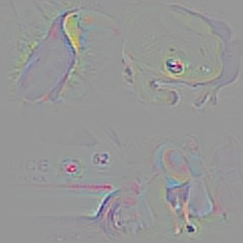} \\
  without pooling & 
  \includegraphics[width=.27\textwidth]{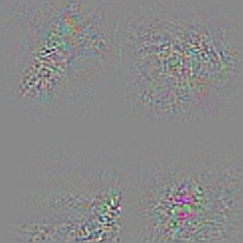} &
  \includegraphics[width=.27\textwidth]{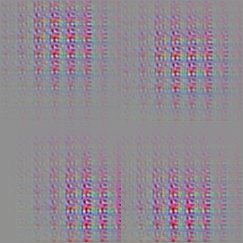} &
  \includegraphics[width=.27\textwidth]{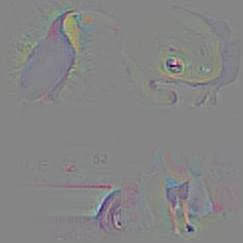} 
\end{tabular}
\end{center}
\caption{Visualization of descriptive image regions with different methods from the single largest activation in the last convolutional layer conv12 of the network trained on ImageNet. Reconstructions for 4 different images are shown.}
\label{fig:cccp8_different_methods}
\end{figure}

\begin{figure}[h]
\begin{center}
\begin{tabular}{ >{\centering\arraybackslash} m{1.5cm} >{\centering\arraybackslash} m{3.5cm} >{\centering\arraybackslash} m{3.5cm} >{\centering\arraybackslash} m{3.5cm} }
   & backpropagation & 'deconvnet' & guided backpropagation \\
  with pooling + switches & 
  \includegraphics[width=.27\textwidth]{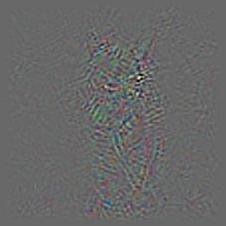} &
  \includegraphics[width=.27\textwidth]{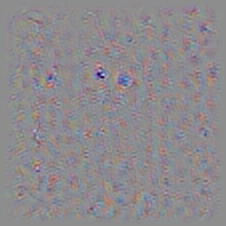} &
  \includegraphics[width=.27\textwidth]{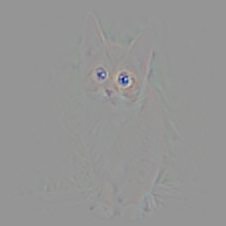} \\
  without pooling & 
  \includegraphics[width=.27\textwidth]{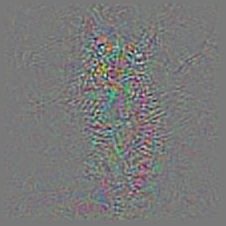} &
  \includegraphics[width=.27\textwidth]{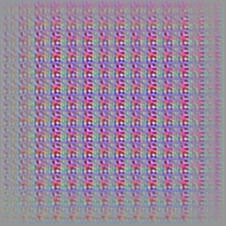} &
  \includegraphics[width=.27\textwidth]{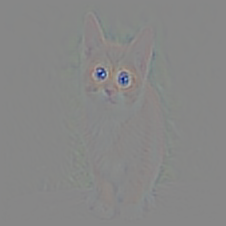} 
\end{tabular}
\end{center}
\caption{Visualization of descriptive image regions with different methods from the single largest activation in the pre-softmax layer global\_pool of the network trained on ImageNet.}
\label{fig:pool4_different_methods}
\end{figure}

\begin{figure}[h]
\begin{center}
\begin{tabular}{c|c|c}
   backpropagation & 'deconvnet' & guided backpropagation \\
  \includegraphics[width=.31\textwidth]{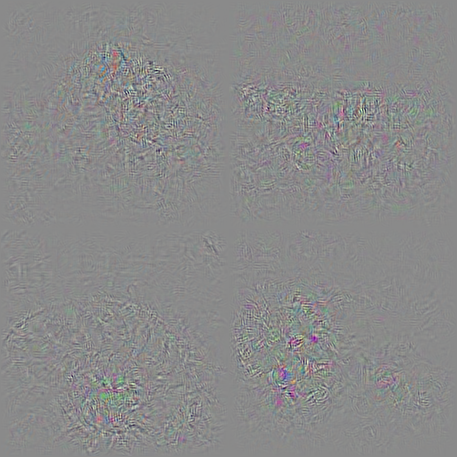} &
  \includegraphics[width=.31\textwidth]{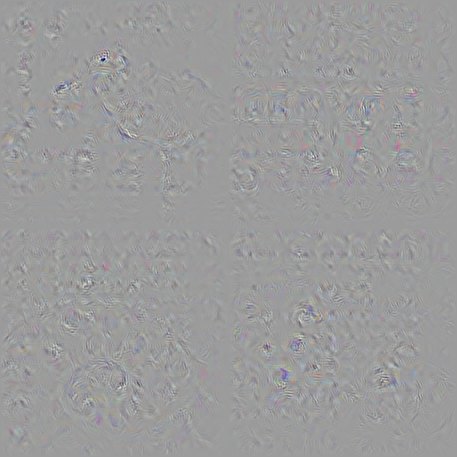} &
  \includegraphics[width=.31\textwidth]{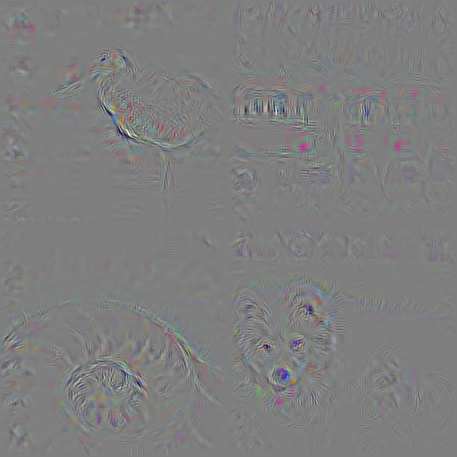} 
\end{tabular}
\end{center}
\caption{Visualization of descriptive image regions with different methods from the single largest activation in the last layer fc8 of the Caffenet reference network~\citep{caffe} trained on ImageNet. Reconstructions for 4 different images are shown.}
\label{fig:fc8_caffenet}
\end{figure}

Additional visualizations of the features learned by the last convolutional layer 'conv12' as well as the pre-softmax layer 'global\_pool' are depicted in Figure \ref{fig:cccp8_different_methods} and Figure \ref{fig:pool4_different_methods} respectively. To allow fair comparison of 'deconvnet' and guided backpropagation, we additionally show in Figure~\ref{fig:fc8_caffenet} visualizations from a model with max-pooling trained on ImageNet.

\end{appendix}

\end{document}